# Conditional Restricted Boltzmann Machines for Structured Output Prediction

**Volodymyr Mnih**
Department of Computer Science
University of Toronto
Toronto, Canada

**Hugo Larochelle** *
Département d'informatique
Université de Sherbrooke
Sherbrooke, Canada

**Geoffrey E. Hinton**
Department of Computer Science
University of Toronto
Toronto, Canada

## Abstract

Conditional Restricted Boltzmann Machines (CRBMs) are rich probabilistic models that have recently been applied to a wide range of problems, including collaborative filtering, classification, and modeling motion capture data. While much progress has been made in training non-conditional RBMs, these algorithms are not applicable to conditional models and there has been almost no work on training and generating predictions from conditional RBMs for structured output problems. We first argue that standard Contrastive Divergence-based learning may not be suitable for training CRBMs. We then identify two distinct types of structured output prediction problems and propose an improved learning algorithm for each. The first problem type is one where the output space has arbitrary structure but the set of likely output configurations is relatively small, such as in multi-label classification. The second problem is one where the output space is arbitrarily structured but where the output space variability is much greater, such as in image denoising or pixel labeling. We show that the new learning algorithms can work much better than Contrastive Divergence on both types of problems.

## 1 Introduction

The number of applications for restricted Boltzmann machines (RBMs) has grown rapidly in the past few years. They have now been applied to multiclass classification (Larochelle & Bengio, 2008), collaborative filtering (Salakhutdinov et al., 2007), motion capture modeling (Taylor & Hinton, 2009), information retrieval (Gehler et al., 2006; Salakhutdinov & Hinton, 2009), modeling natural images (Osindero & Hinton, 2008) and many other tasks. One problem that has not received much attention in the RBM literature is structured output prediction. In this paper, we are particularly interested in structured output prediction problems where no obvious and reasonable simplifying assumptions can be made about the interactions between the outputs, such as chain-like interactions in sequential classification problems. One such problem is multi-label classification, where examples can be labeled as simultaneously belonging to several classes. Another broad class of problems includes image labeling, where the goal is to assign a discrete label to each pixel or region of an image based on some features extracted from the image.

One difficulty in structured output prediction is that the output space typically has an exponential number of possible configurations. In the case of training RBMs, such a large number of output configurations means that exact gradients are intractable and approximations to the gradient such as Contrastive Divergence (CD) learning need to be used (Hinton, 2002). While a number of improved algorithms for training RBMs with large output spaces have recently been developed, as we will show, none of them apply to the problem of training conditional RBMs. In fact, for large output spaces where exact gradients are intractable (Larochelle & Bengio, 2008), CD learning has been the only algorithm used to train CRBMs (Salakhutdinov et al., 2007; Taylor & Hinton, 2009; Memisevic & Hinton, 2010).

In this work, we argue that CD learning may not be a very good algorithm for training CRBMs and propose two new algorithms for tackling structured output prediction problems in two different settings. In the first, we can assume that the variability in the output space is limited, meaning that the available training data covers the set of likely output configurations relatively well. One example of such problem is multi-label classification. We propose to use semantic hashing in order to define and efficiently compute, a small set of possible outputs to predict given some input. This allows us to perform exact inference over this set, under the CRBM's energy function.

---
* This work was done while Hugo was at University of Toronto

In the second setting, we simply assume that the output space is high-dimensional and highly variable. Denoising and image labeling are problems that fall in this category. In this context, we propose a perceptron-like algorithm for training conditional RBMs. We then demonstrate that CD-based training of a conditional RBM fails to find a good solution on a denoising problem while perceptron-based training succeeds.

## 2 Training RBMs

We begin with an overview of maximum likelihood learning in RBMs before proceeding to learning in conditional RBMs.

### 2.1 Restricted Boltzmann Machines

A Restricted Boltzmann Machine is an undirected graphical model that defines a probability distribution over a vector of observed, or *visible*, variables $\mathbf{v}$ and a vector of latent, or *hidden*, variables $\mathbf{h}$. In this paper, we consider the case where $\mathbf{v}$ and $\mathbf{h}$ are binary vectors. An RBM defines a joint probability over $\mathbf{v}$ and $\mathbf{h}$,

$$p(\mathbf{v},\mathbf{h}) = \exp\left(-E\left(\mathbf{v},\mathbf{h}\right)\right)/Z, \qquad (1)$$

where $Z$ is a normalization constant and $E$ is an energy function given by

$$E\left(\mathbf{v},\mathbf{h}\right) = -\mathbf{v}^T\mathbf{W}\mathbf{h} - \mathbf{v}^T\mathbf{b}^v - \mathbf{h}^T\mathbf{b}^h. \qquad (2)$$

$\mathbf{W}$ is a matrix of pairwise weights between elements of $\mathbf{v}$ and $\mathbf{h}$, while $\mathbf{b}^v$ and $\mathbf{b}^h$ are biases for the visible and hidden variables respectively. To obtain $p(\mathbf{v})$ one simply marginalizes out $\mathbf{h}$ from the joint distribution:

$$p(\mathbf{v}) = \sum_{\mathbf{h}} \exp\left(-E(\mathbf{v},\mathbf{h})\right)/Z = \exp\left(-F(\mathbf{v})\right)/Z \quad (3)$$

where $F(\mathbf{v})$ is called the free energy and can be computed in time linear in the number of elements in $\mathbf{v}$ and $\mathbf{h}$:

$$F(\mathbf{v}) = -\log \sum_{\mathbf{h}} \exp\left(-E(\mathbf{v},\mathbf{h})\right) \qquad (4)$$

$$= -\mathbf{v}^T\mathbf{b}^v - \sum_j \log\left(1 + \exp\left(b_j^h + \mathbf{v}^T\mathbf{W}_{\cdot j}\right)\right) \qquad (5)$$

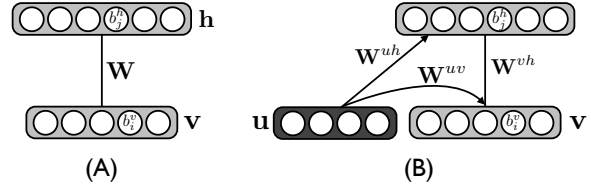

Figure 1: Illustration of an RBM (A) and a conditional RBM (B).

RBMs have generally been trained using gradient descent in negative log-likelihood $-l(\theta)$ for some set of training vectors $\mathbf{V}$. By writing the log-likelihood as

$$\log p(\mathbf{v}) = \log \exp\left(-F(\mathbf{v})\right) - \log \sum_{\mathbf{v}'} \exp\left(-F(\mathbf{v}',)\right), \qquad (6)$$

and differentiating $-l(\theta)$ with respect to some parameter $\theta$, we get the gradient

$$\frac{\partial -l(\theta)}{\partial \theta} = \frac{\partial F(\mathbf{v})}{\partial \theta} - \sum_{\mathbf{v}'} \frac{\partial F(\mathbf{v}')}{\partial \theta} p(\mathbf{v}'). \qquad (7)$$

The first term in Equation 7 can be computed exactly. This term is often referred to as the *positive* gradient. It also corresponds to the expected gradient of the energy (as opposed to the free energy), where the expectation is with respect to $p(\mathbf{h}|\mathbf{v})$. This simplification occurs because the gradient of $F$ w.r.t. $p(\mathbf{h}|\mathbf{v})$ is zero, so the effect of changing the parameters on $p(\mathbf{h}|\mathbf{v})$ can be ignored.

The second term in Equation 7, known as the *negative* gradient, is an expectation over the model distribution, $p(\mathbf{v})$, and is intractable to compute exactly for all but the smallest models. It is possible to estimate the negative gradient by drawing samples from the model using MCMC methods. Since both $p(\mathbf{v}|\mathbf{h})$ and $p(\mathbf{h}|\mathbf{v})$ factor over the variables, it is possible to efficiently perform Gibbs sampling by alternating between updating all of $\mathbf{v}$ and all of $\mathbf{h}$ simultaneously. We can then ignore the sampled $\mathbf{h}$ and only keep the sampled $\mathbf{v}$. Nevertheless, running a Gibbs chain until equilibrium for each parameter update is not feasible.

### 2.2 Contrastive Divergence

The first practical method for training RBMs was introduced by Hinton (2002), who showed that the negative gradient can be approximated using samples obtained by starting a Gibbs chain at a training vector and running it for a few steps. This method approximately minimizes an objective function known as the Contrastive Divergence. Even though it has been shown that the resulting gradient estimate is not the gradient of any function (Sutskever & Tieleman, 2010), CD learning has been used extensively for training RBMs and other energy-based models.

### 2.3 Persistent Contrastive Divergence

One problem with CD learning is that it provides biased estimates of the gradient. The Persistent Contrastive Divergence (PCD) algorithm (Tieleman, 2008) addressed this problem. In the positive phase, PCD does not differ from CD training. In the negative phase, however, instead of running a new chain for each parameter update, PCD maintains a single *persistent* chain. The update at time $t$ takes the state of the Gibbs chain at time $t-1$, performs one round of Gibbs sampling, and uses this state in the negative gradient estimates. When the learning rate is small, the model does not change much between updates and the state of the persistent chain is able to stay close to the model distribution, leading to more accurate gradient estimates.

A number of other algorithms have been proposed for obtaining better gradient estimates for training RBMs (Desjardins et al., 2010; Salakhutdinov, 2010; Tieleman & Hinton, 2009). However, these algorithms make use of persistent Markov chains for obtaining improved gradient estimates and, as we will later show, this makes these algorithms, as well as PCD, not applicable to training of conditional RBMs.

## 3 Training Conditional RBMs

A CRBM (see figure 1) models the distribution $p(\mathbf{v}|\mathbf{u})$ by using an RBM to model $\mathbf{v}$ and using $\mathbf{u}$ to dynamically determine the biases or weights of that RBM. In this paper we only allow $\mathbf{u}$ to determine increments to the visible and hidden biases of the RBM:

$$E(\mathbf{v}, \mathbf{h}, \mathbf{u}) = -\mathbf{v}^T \mathbf{W}^{vh} \mathbf{h} - \mathbf{v}^T \mathbf{b}^v - \mathbf{u}^T \mathbf{W}^{uv} \mathbf{v} \\ - \mathbf{u}^T \mathbf{W}^{uh} \mathbf{h} - \mathbf{h}^T \mathbf{b}^h \quad (8)$$

with the associated free energy

$$F(\mathbf{v}, \mathbf{u}) = -\log \sum_{\mathbf{h}} \exp(-E(\mathbf{v}, \mathbf{h}, \mathbf{u})) \quad (9)$$

$$= -\sum_j \log\left(1 + \exp\left(b_j^h + \mathbf{v}^T \mathbf{W}^{vh}_{\cdot j} + \mathbf{u}^T \mathbf{W}^{uh}_{\cdot j}\right)\right) \\ - \mathbf{v}^T \mathbf{b}^v - \mathbf{u}^T \mathbf{W}^{uv} \mathbf{v}. \quad (10)$$

Note that the algorithms we propose in this paper are not restricted to this model and can be used to train any energy-based model for which the free-energy (or just the energy, for models without latent variables) can be computed. For the CD-PercLoss algorithm there is an additional requirement that Gibbs sampling can be performed.

The CRBM model defines the following probability distribution:

$$p(\mathbf{v}|\mathbf{u}) = \frac{\exp(-F(\mathbf{v}, \mathbf{u}))}{\sum_{\mathbf{v}'} \exp(-F(\mathbf{v}', \mathbf{u}))}. \quad (11)$$

Learning in conditional RBMs generally involves doing gradient descent in negative log conditional likelihood. The gradient of the negative log conditional likelihood for a conditional RBM is given by

$$\frac{\partial -l(\theta)}{\partial \theta} = \frac{\partial F(\mathbf{v}|\mathbf{u})}{\partial \theta} - \sum_{\mathbf{v}'} \frac{\partial F(\mathbf{v}', \mathbf{u})}{\partial \theta} p(\mathbf{v}'|\mathbf{u}). \quad (12)$$

In some cases this gradient can be computed exactly (Larochelle & Bengio, 2008), but it is intractable in general. However, since $p(\mathbf{v}|\mathbf{h}, \mathbf{u})$ and $p(\mathbf{h}|\mathbf{v}, \mathbf{u})$ are both factorial over $\mathbf{v}$ and $\mathbf{h}$, CD can be used to train a conditional RBM – the positive gradient is still tractable and it is still possible to do block Gibbs sampling to approximate the negative gradient.

Almost all algorithms that have been proposed for training RBMs are not applicable to training conditional RBMs because they make use of persistent chains. With a conditional RBM, each conditioning vector $\mathbf{u}$ generally leads to a unique distribution $p(\mathbf{v}|\mathbf{u})$, hence one needs to sample from a different model distribution for each training case, making it impossible to use a single persistent chain. One could run a separate persistent chain for each training case, but this is only feasible on very small datasets. To learn efficiently on large datasets we need to update the weights on mini-batches of training data that are much smaller than the whole training set, so by the time we revisit a training case the weights will have changed substantially and the persistent chain for that case will be far from its stationary distribution.

Unlike algorithms based on persistent chains, a number of the algorithms recently proposed for training non-normalized statistical models could be applied to conditional RBM training (Hyvärinen, 2007; Vickrey et al., 2010). While we only compare our approaches to algorithms that have been applied to conditional RBMs, the Contrastive Constraint Generation (CCG) algorithm (Vickrey et al., 2010) is the most closely related approach to our work. CCG is a batch algorithm that maintains a contrastive set of points for each training case, where the contrastive points are generated by running some inference procedure. Both of our proposed approaches have similarities to the CCG algorithm. The HashCRBM algorithm uses spectral hashing to define the contrastive neighborhoods and only includes values of $\mathbf{v}$ that are in the training set, which ensures that the total number of unique contrastive points is reasonable. The CD-PercLoss algorithm generates contrastive examples using inference, but it is an online algorithm, and hence can scale up to large datasets unlike CCG.

## 4 Making Predictions with Conditional RBMs

While RBMs have primarily been used for learning new representations of data, conditional RBMs are generally used for making predictions. This typically requires inferring either the modes of the conditional marginals $p(v_i|\mathbf{u})$ or the mode of the full conditional $p(\mathbf{v}|\mathbf{u})$. Such inferences are intractable in general, hence approximate inference algorithms must be used.

The most popular in the RBM literature for conditional models is mean-field inference (Salakhutdinov et al., 2007; Mandel et al., 2011), which is a fast message passing algorithm and can work well in practice. In the context of structured prediction, it assumes a full conditional that factorizes into its marginals $q(\mathbf{v}|\mathbf{u}) = \prod_i q(v_i|\mathbf{u})$ and finds the marginals that provide the best approximation of the true conditional distribution.

Loopy belief propagation (Murphy et al., 1999) is another possible message passing algorithm that could be used and that tends to provide better approximations of the conditional marginals of CRBMs (Mandel et al., 2011). Unfortunately, it is also slower than mean-field, and we have found it to be practical only on problems with relatively small output dimensionality and hidden layer size.

## 5 Suitability of CD Training

In this section we argue that CD training may not be very well suited to training conditional RBMs. One way to look at CD training of RBMs is as a process that lowers the free energy of the data $\mathbf{v}$ in the positive phase and raises the free energy of the $k$th state of a Gibbs chain started at the data $\mathbf{v}$. Such a process can be seen as minimizing the following loss function

$$L_{CD}(\mathbf{v}, \mathbf{u}|\theta) = F(\mathbf{v}, \mathbf{u}|\theta) - F(\mathbf{v}_k(\theta_{\text{old}}), \mathbf{u}|\theta), \quad (13)$$

where $\mathbf{v}_k(\theta_{\text{old}})$ is the $k$th state of a Gibbs chain that is started at the data and uses parameters $\theta_{\text{old}}$ to determine the transition matrix. When the parameters $\theta$ are updated, the Markov chain used to produce $\mathbf{v}_k$ also changes, but the CD learning procedure ignores the effect this has on the second free energy term in Eq 13.

If the negative samples $\mathbf{v}_k$ have the same distribution as the data $\mathbf{v}$, the loss function $L_{CD}$ will be zero on average, meaning that we have successfully learned the data distribution $p(\mathbf{v}|\mathbf{u})$. However, since the negative phase Gibbs chain starts at the data, this objective function will also be small if the chain is mixing slowly. Indeed, if the chain starts at $\mathbf{v}$ and is mixing slowly then, for small $k$, $\mathbf{v}_k$ will be very close to $\mathbf{v}$, in turn making $L_{CD}$ small. This is particularly troubling because, when training a conditional RBM, the aim is usually to be able to make good predictions for $\mathbf{v}$ when given a conditioning vector $\mathbf{u}$ and making $L_{CD}$ small gives no guarantees about the quality of the predictions.

It is interesting to note that CD training of non-conditional RBMs also suffers from the same problem, i.e. the gradient estimate will be small when the negative phase Gibbs chain is mixing slowly. Nevertheless, this is not as much of an issue when training non-conditional RBMs because they are generally used for learning representations of the data. If a Gibbs chain started at the data $\mathbf{v}$ is mixing slowly and not moving far from $\mathbf{v}$, then the hidden representation $\mathbf{h}$ must contain enough information to reconstruct $\mathbf{v}$ well, possibly making $\mathbf{h}$ a reasonable representation of $\mathbf{v}$. In a conditional RBM being able to reconstruct the data $\mathbf{v}$ is not sufficient for making good predictions, because at prediction time one only has access to $\mathbf{u}$ and not $\mathbf{v}$.

Therefore, it seems that while CD is a reasonable training procedure for non-conditional RBMs, it may not be a reasonable training procedure for conditional RBMs because it does not directly encourage the model to make good predictions. We provide experimental support for this argument by showing that CD training of conditional RBMs can fail on seemingly simple structured output prediction problems.

## 6 Two Algorithms for Structured Output Prediction with CRBMs

We propose two alternatives to CD training of CRBMs. The first alternative is appropriate when the variability of observed configurations for $\mathbf{v}$ is limited and the output space is relatively well covered by the available training data. The second looks at the case where such assumptions are not reasonable and tries to address the structured output problem in general.

### 6.1 Structured Output Prediction when Output Variability is Limited

As mentioned before, the intractability of learning and inference is directly related to the exponential size of the output space – if the number of possible configurations for the output target $\mathbf{v}$ was small enough, an exhaustive enumeration of these configurations would be possible and exact training/inference could be considered.

Part of the reason for the exponential size of **v** is that we wish to be conservative with respect to the potential configurations of **v** that the model can output and not discard any of them before training. For certain problems however, being this conservative is not necessary. In particular, if there are enough available training data to cover (or come very close to) all reasonable configurations of **v**, it would be sufficient to constrain the predictions to correspond to a value of **v** previously observed. In other words, we would define

$$p(\mathbf{v}|\mathbf{u}) = \frac{1_{\mathbf{v} \in \mathcal{V}} \exp\left(-F(\mathbf{v}, \mathbf{u})\right)}{\sum_{\mathbf{v}' \in \mathcal{V}} \exp\left(-F(\mathbf{v}', \mathbf{u})\right)} \qquad (14)$$

where the difference with Equation 11 is that we have explicitly set to zero the probability of outputs not in the set $\mathcal{V} \subseteq \{0,1\}^{|\mathbf{v}|}$, corresponding to all values of **v** observed in the training set.

### Spectral Hashing

While technically possible, this approach will be terribly slow for large $\mathcal{V}$, even if relatively small compared to $\{0,1\}^{|\mathbf{v}|}$. Moreover, given some input **u**, only a subset of $\mathcal{V}$ will correspond to likely prediction candidates. So computationally, it would be even better to construct a subset $\mathcal{V}(\mathbf{u}) \subseteq \mathcal{V}$ for each given input **u**, as long as constructing this subset is fast.

Fortunately, this can be achieved by viewing the problem of constructing this subset $\mathcal{V}(\mathbf{u})$ as one of performing fast retrieval of relevant "documents" **v** for some given "query" **u**. The spectral hashing algorithm of Weiss et al. (2009) then provides a method for constructing such subsets without having to explicitly search through the training set.

The original spectral hashing algorithm allows for fast retrieval of (approximate) nearest neighbors by learning a binary code for the inputs **u** and then constructing a hash map where the values are the training set inputs **u** and the keys are their associated binary codes. Neighbors can then be retrieved by computing the given input's binary code and using the hash map to fetch the training inputs whose binary codes are within a small hamming distance. In this paper we use a maximum hamming distance of 1 which gives $n+1$ hash map accesses for a binary code of $n$ bits. We could probably achieve better results at the cost of more computation by using a slightly larger hamming ball, especially for codes with many bits or for codes that return few candidates because their immediate neighbors never occur in the training data.

One important advantage of spectral hashing over other variants like its precursor semantic hashing (Salakhutdinov & Hinton, 2007) is that training is very efficient and has an analytical solution. It is based on a spectral relaxation of the general semantic hashing problem and mainly requires the computation of a few principal components (PCA) of the inputs. See Weiss et al. (2009) for more details.

In the context of this work, there is one simple modification we must apply to the original spectral hashing algorithm. Since we wish to obtain a set of outputs $\mathcal{V}(\mathbf{u})$, when constructing the hash map, we associate to the binary code keys the training set target **v** associated with the input **u** that generated the binary code. Moreover, when merging the $n+1$ sets of outputs **v** obtained from each hash map access, duplicated configurations of **v** must be filtered out[1].

### Training and Inference

When using input dependent output subsets $\mathcal{V}(\mathbf{u})$, the CRBM conditional distribution becomes:

$$p(\mathbf{v}|\mathbf{u}) = \frac{1_{\mathbf{v} \in \mathcal{V}(\mathbf{u})} \exp\left(-F(\mathbf{v}, \mathbf{u})\right)}{\sum_{\mathbf{v}' \in \mathcal{V}(\mathbf{u})} \exp\left(-F(\mathbf{v}', \mathbf{u})\right)} \qquad (15)$$

where we now explicitly assign a probability of zero to even more output configurations. We can then train this CRBM by **exact** gradient descent on the negative log conditional likelihood. The gradient is similar to that of Equation 12, but where the so-called negative gradient requires a sum over elements in $\mathcal{V}(\mathbf{u})$ only. No approximations are needed. We will refer to CRBMs trained using this proposed spectral hashing method as HashCRBMs.

For inferring a prediction at test time, one simple option would be to output the element of $\mathcal{V}(\mathbf{u})$ with smallest free energy $F(\mathbf{v}, \mathbf{u})$, which would correspond to predicting the mode of the global conditional $p(\mathbf{v}|\mathbf{u})$. However, the quality of this prediction relies heavily on the quality of potential predictions in $\mathcal{V}(\mathbf{u})$. In particular, with predictions based on the conditional mode, if the correct target isn't present in $\mathcal{V}(\mathbf{u})$, then there is no way for the HashCRBM to make a perfect prediction.

Another option would be to make predictions for each element of **v** based on the modes of each marginal $p(v_i|\mathbf{u})$. Noting $\hat{\mathbf{v}}$ as the HashCRBM prediction, this would correspond to computing

$$\hat{v}_i = \underset{a \in \{0,1\}}{\operatorname{argmax}} p(v_i = a | \mathbf{u}) = \underset{a \in \{0,1\}}{\operatorname{argmax}} \sum_{\substack{\mathbf{v} \in \mathcal{V}(\mathbf{u}) \\ \text{s.t. } v_i = a}} p(\mathbf{v}|\mathbf{u}). \qquad (16)$$

---

[1]This issue doesn't come up in the original spectral hashing since the hash map values directly generated the keys (binary codes), and hence the same hash map value can only be associated with one key

Such a prediction is much more appropriate when the performance measure for the task to solve decomposes into individual costs for each element of $\hat{\mathbf{v}}$. More importantly, this allows the HashCRBM to make predictions that are not in $\mathcal{V}(\mathbf{u})$. For instance, if $\mathcal{V}(\mathbf{u}) = \{[1,1,0], [1,0,1], [0,1,1]\}$ and the free energy $F(\mathbf{v}, \mathbf{u})$ is the same for all $\mathbf{v} \in \mathcal{V}(\mathbf{u})$, then the HashCRBM prediction will be $\hat{v} = [1,1,1] \notin \mathcal{V}(\mathbf{u})$.

#### 6.1.1 Experiments with HashCRBM

In multi-label classification, the target $\mathbf{v}$ to predict corresponds to a binary vector which encodes whether input $\mathbf{u}$ belongs to class $l$ ($u_l = 1$) or not ($u_l = 0$), for all classes $l \in \{1, \ldots, L\}$. Alternatively, we say that $\mathbf{u}$ has label $l$ if $u_l = 1$. Labels (or classes) are not mutually exclusive, hence the name multi-label classification. Modeling the tendency of certain labels to co-occur can then bring potential improvements in performance.

We compare the HashCRBM of Section 6.1 to three other different models: a model consisting of several label-specific logistic regressors (LogReg), a conditional random field (CRF) with fully connected output units and a standard CRBM trained using contrastive divergence. We looked at either mean-field (CRBM + MF) or loopy belief propagation (CRBM + LBP) to perform approximate inference of the conditional marginals $p(v_i|\mathbf{u})$, with different numbers of iterations (in $\{5, 10, 20\}$). For the CRF, we used 50 iterations of loopy belief propagation to approximate the conditional likelihood gradients and to make predictions. Whenever loopy belief propagation was used, different damping factors (in $\{0, 0.3, 0.6\}$) were also tested. All models made predictions at test time based on the (estimated) conditional marginals' modes, which is appropriate for multi-label classification since the performance measure is the average of the individual label classification errors.

All models were trained by stochastic gradient descent, with learning rate chosen among $\{2^{-4}, 2^{-6}, 2^{-8}, 2^{-10}\}$. Hidden layer sizes in $\{32, 64, 128, 256\}$, number of Gibbs steps for CD training in $\{1, 10, 20\}$ were tested for both CRBM baselines. For HashCRBM, binary codes of size $n$ in $\{5, 7, 9\}$ were used. The best combination of hyper-parameters values was selected based on the validation set performance for each model. Early stopping based on the validation set error progression was also employed.

Table 1 gives the average (over all labels) classification error for each model, on four different datasets: Yeast (Elisseeff & Weston, 2002), Scene (Boutell et al., 2004), MTurk (Mandel et al., 2010) and MajMin (Mandel & Ellis, 2008). The first two are com-mon public benchmarks for multi-label classification. The last two correspond to music tagging problems and were provided by the authors of Mandel et al. (2010) and Mandel and Ellis (2008). For each dataset, we report the average test performance over 10 folds, each consisting in a training (80%), validation (10%) and test (10%) split of the whole data.

| Model | YEAST | SCENE | MTURK | MAJMIN |
|---|---|---|---|---|
| LogReg | **19.92** | 10.55 | 7.32 | 4.26 |
| CRF | 21.03 | 9.75 | **7.21** | **4.24** |
| CRBM + MF | 20.73 | 9.85 | **7.27** | 4.28 |
| CRBM + LBP | 20.58 | 9.47 | **7.25** | **4.22** |
| HashCRBM | **20.02** | **8.80** | **7.24** | **4.20** |

Table 1: Average test errors for multi-label experiment on four datasets. For each dataset, we put in bold the lowest average error as well as any other average error that is not statistically significantly different from the lowest, based on a 95% two-sided t-test on the test error differences for the 10 folds.

As we can see, the HashCRBM is consistently amongst the best performing models, with a particularly big improvement over all methods on the Scene dataset.

### 6.2 General Structured Output Prediction

**Minimizing the Generalized Perceptron Loss**

As we argued in Section 5, Contrastive Divergence learning may not be a suitable procedure for training conditional RBMs because it does not seem to directly discourage the model from making bad predictions. We propose circumventing this problem in training conditional RBMs by instead minimizing a loss function that depends on the model's prediction. Given a training case $(\mathbf{v}, \mathbf{u})$, the generalized perceptron loss (LeCun et al., 2006) is defined as

$$L_P(\mathbf{v}, \mathbf{u}, \theta) = F(\mathbf{v}, \mathbf{u}, \theta) - \min_{\mathbf{v}^*(\theta_{\text{old}})} F(\mathbf{v}^*(\theta_{\text{old}}), \mathbf{u}, \theta). \quad (17)$$

When using $\operatorname{argmin}_{\mathbf{v}^*} F(\mathbf{v}^*, \mathbf{u})$ as the model's prediction for a vector $\mathbf{u}$, minimizing the generalized perceptron loss will directly discourage the model from making bad predictions by continuously raising the free energy of the model's predictions and lowering the free energy of the data.

Since the minimization in Equation 17 is equivalent to finding the mode of the distribution $p(\mathbf{v}|\mathbf{u})$, it is, in general, intractable for conditional RBMs. When this minimization cannot be performed exactly, we propose using the free energy at the model's prediction $\hat{\mathbf{v}}$ for the conditioning vector $\mathbf{u}$ in place of the minimum free energy value. The loss function $L_P$ can then be ap-proximately optimized efficiently for the same class of energy functions for which the free energy can be computed efficiently. Intuitively, training a conditional RBM in this manner has the appealing quality that the gradient should be large when the model makes a bad prediction.

To fully specify our training algorithm we now describe our procedure for making predictions. First, we initialize the states of the visible units. When connections between $\mathbf{u}$ and $\mathbf{v}$ are present, we initialize $\mathbf{v}_i$ to $\sigma(b_i^v + \mathbf{u}^T \mathbf{W}_{\cdot i}^{uv})$, where $\sigma(x) = 1/(1 + \exp(-x))$. These starting values for $\mathbf{v}_i$ correspond to the probabilities given by the logistic regression component of the conditional RBM. When there are no connections between the conditioning vector $\mathbf{u}$ and the visible units $\mathbf{v}$ we initialize visible units randomly by setting each unit to 1 with probability 0.5 and 0 otherwise. Starting with the initialized states for $\mathbf{v}$ we run $k$ steps of Gibbs sampling producing $k$ sets of states of the visible units $\mathbf{v}^{(1)}, \ldots, \mathbf{v}^{(k)}$ (one for each iteration of Gibbs sampling). Finally, out of these $k$ configurations, we select the configuration of the visible units with the lowest free energy as the prediction. This procedure corresponds to a stochastic search through the space of possible outputs.

The proposed training procedure then corresponds to minimizing

$$L_P(\mathbf{v}, \mathbf{u}) = F(\mathbf{v}, \mathbf{u}) - F(\hat{\mathbf{v}}, \mathbf{u}),$$

where $\hat{\mathbf{v}}$ is seen as a constant and is the prediction produced by the stochastic search procedure (in other words we ignore the dependence of $\hat{\mathbf{v}}$ on the parameters during optimization). Since stochasticity is not a desirable property of a prediction procedure, at test time, we use $k$ steps of mean field inference instead of $k$ steps of Gibbs sampling to produce $\mathbf{v}^{(1)}, \ldots, \mathbf{v}^{(k)}$. We refer to this algorithm as CD-PercLoss.

### 6.2.1 Experiments

We consider the problem of image denoising, where the goal is to predict a clean image $\mathbf{v}$ from some noisy image $\mathbf{u}$. We use two datasets derived from the MNIST digit database[2]. In both cases, we first binarized all digits by thresholding the pixel values at 0.5.

In the first dataset, which we will refer to as *corrupted MNIST*, the vector $\mathbf{v}$ is the clean binarized digit while the vector $\mathbf{u}$ is obtained by flipping 10% of the entries in $\mathbf{v}$. Having a good model of digits is not essential for doing well on this dataset because it is possible to do well just by doing local voting among entries of $\mathbf{u}$ when making a prediction.

---

[2] We do not provide results on the multi-label classification datasets used to evaluate HashCRBM because we do not expect it to outperform HashCRBM in this setting, and preliminary results confirm that CD-PercLoss has performance comparable to CRBM+MF on such problems.

For the second dataset, the vector $\mathbf{u}$ is obtained by setting a random 8 by 8 patch of the image $\mathbf{v}$ to 0. With this type of noise, the distribution $p(\mathbf{v}|\mathbf{u})$ can be bimodal when, for example, the upper part of a 7 is replace by 0s, making it difficult to determine whether the original digit was a 1 or a 7. An approach that takes structure in the entries of $\mathbf{v}$ into account should do well on this type of problem. We will refer to this dataset as *occluded MNIST*.

CRBMs with the energy function of Equation 8 should be able to do well on both the corrupted and occluded MNIST tasks. The weights $\mathbf{W}$ should learn features that are good for representing handwritten digits, such as parts of digits. The weights $\mathbf{W}^{uh}$ then correspond to a set of linear filters on the vector $\mathbf{u}$ (one for each hidden unit) that bias the hidden units to being on or off. The weights $\mathbf{W}^{uv}$ similarly bias visible units.

We fix the number of hidden units at 256 and train the model with CD-$k$ for $k = 1$ and $k = 10$ for 128 epochs. We also test logistic regression and the CD-PercLoss algorithm with 10-step predictions. We used stochastic gradient descent with batches of size 128 and did not use momentum or weight decay. A fixed learning rate was selected from the choices in $\{2^0, 2^{-2}, 2^{-4}, \ldots, 2^{-14}\}$ using a validation set. Early stopping based on the validation error was also used.

Table 2 shows the fraction of incorrectly labeled pixels on the corrupted and occluded MNIST for the different algorithms. We also include a baseline that uses the corrupted image as the prediction to show that the algorithms are indeed cleaning up the noisy images. As expected, all algorithms perform well on the Corrupted MNIST task with the CD-PercLoss algorithm slightly outperforming the others.

On the Occluded MNIST dataset, the CD-PercLoss training algorithm outperforms the other algorithms by a wider margin. The error rate on the pixels that were deleted by the noise process for the model trained with CD-PercLoss is 20% lower than the models trained with CD. Somewhat surprisingly, even logistic regression achieves lower error than the conditional RBMs trained with CD-1 and CD-10. Since the weights $\mathbf{W}^{uv}$ essentially define a logistic regression component inside the conditional RBM model, the inability of CD to train a richer model to achieve lower error on a dataset where structured prediction should help confirms our theory that CD may not be a good algorithm for training such models. Figure 2 shows the predictions for the best models trained with CD-1, CD-10 and CD-PercLoss. This figure shows that while the error rates over all pixels seem quite low, the models trained with CD-1 and CD-10 are unable to reconstruct most digits correctly while the model trained with CD-PercLoss correctly cleans up most digits.

| Dataset | Corrupted | | Occluded | |
| Model | All | Changed | All | Changed |
| --- | --- | --- | --- | --- |
| Baseline | 9.993 | 100.0 | 1.920 | 100.0 |
| LogReg | 1.937 | 12.54 | 1.560 | 60.10 |
| CD-1 | 1.923 | 10.95 | 1.800 | 62.90 |
| CD-10 | 1.816 | 11.1 | 1.704 | 67.93 |
| CD-PercLoss | **1.755** | **10.71** | **1.357** | **41.89** |

Table 2: Average test errors for the denoising experiment on MNIST. Two numbers are shown for each model and dataset. All Denotes the percentage of incorrectly labeled pixels among all pixels. Changed denotes the percentage of incorrectly labeled pixels among pixels that were changed by the noise process. Bold has the same meaning as in Table 1.

## 7 Conclusions and Future Work

We have presented two new algorithms for training conditional RBMs and shown that they perform better than alternative methods in the regimes for which they are appropriate. The CD-PercLoss procedure is actually more like the original learning procedure for Boltzmann Machines than like contrastive divergence because the chain is started at a random state instead of the data.

The training procedure that was first proposed for Boltzmann machines (Ackley et al., 1985) involved estimating the derivative of the log partition function by starting both the hidden and the visible units at a random state and then using simulated annealing down to a temperature of 1 to try to sample from the stationary distribution. For RBMs, contrastive divergence was a practical advance on the original learning procedure for two main reasons. First, if the model is highly multi-modal with widely separated modes, it is hard to ensure that randomly initiated chains sample the different modes with the right relative frequencies. For a good model, the modes of the model should roughly correspond to the modes of the data distribution, so starting the chain at a random datapoint helps to ensure that it samples the modes better, especially towards the end of learning. Second, when training on large datasets using small mini-batches, starting the chain at the data helps to ensure that the sampling error in the data caused by using a small mini-batch is roughly balanced by a highly correlated sampling error in the "negative" data.

For CRBMs, the joint distribution of the visible and hidden units conditioned on $u$ is generally much less multi-modal than the joint distribution in an RBM, so the advantages of CD over the original learning procedure can be diminished to the point where it is no longer an effective technique.

One interesting direction of future work is to investigate a structured output max-margin variant of CD-PercLoss, with loss margin scaling. This should only require two changes: that loss-augmented inference be performed by subtracting a loss term $l(\mathbf{v}^*(\theta_{\text{old}}), \mathbf{v})$ inside of the minimization of Equation 17, and that a parameter update be performed only when the difference in free energies of $\hat{\mathbf{v}}$ and $\mathbf{v}$ is smaller than the loss $l(\hat{\mathbf{v}}, \mathbf{v})$. Using the hamming loss, it should be straightforward to define loss-adapted Gibbs sampling iterations for the approximate stochastic minimization search. L2 regularization would also need to be added. HashCRBM could similarly be adjusted.

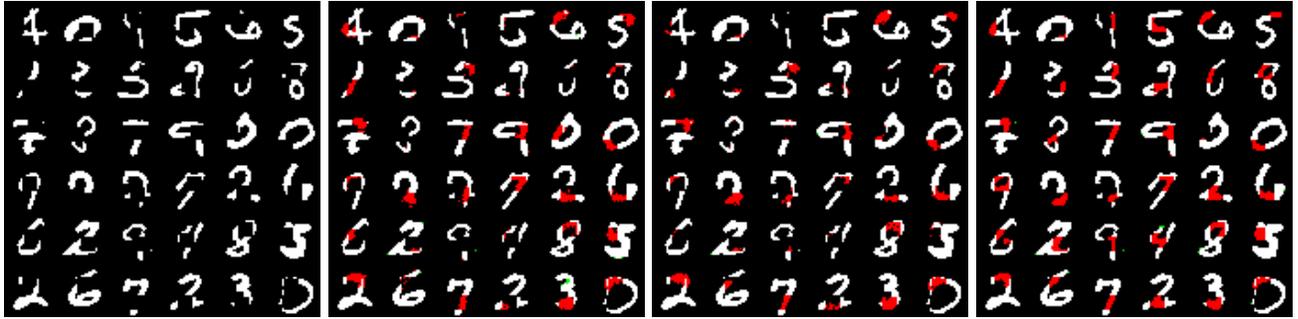

Figure 2: Best viewed in color. From left to right: 1) Noisy images **u**. 2) Predictions for the best model trained with CD-1 3) Predictions for the best model trained with CD-10 4) Predictions for the best model trained with CD-PercLoss. Pixels present in both the noisy and the predicted images are shown in white. Pixels present in the predicted images but not the noisy images are shown in red.